\documentclass[sigconf, nonacm]{acmart}
\usepackage{algorithm}
\usepackage[noend]{algpseudocode}
\usepackage{array}
\begin{document}

\title{Temporal Leakage Distorts the TGB Node Affinity Leaderboard: A Point-in-Time Audit}

\author{Yuhang Wang}
\affiliation{%
	\institution{Beijing Forestry University}
	\city{Beijing}
	\country{China}
}
\email{yuhang1103@bjfu.edu.cn}

\author{Dongmei Li}
\authornote{Corresponding authors.}
\affiliation{%
	\institution{Beijing Forestry University}
	\city{Beijing}
	\country{China}
}
\email{lidongmei@bjfu.edu.cn}

\author{Yuze Sun}
\authornotemark[2]
\affiliation{%
	\institution{Tsinghua University}
	\city{Beijing}
	\country{China}
}
\email{syz23@mails.tsinghua.edu.cn}

\author{Xiaolong Qu}
\affiliation{%
	\institution{Beijing Forestry University}
	\city{Beijing}
	\country{China}
}
\email{quxiaolong@bjfu.edu.cn}

\author{Yanxiang Wang}
\affiliation{%
	\institution{Beijing Forestry University}
	\city{Beijing}
	\country{China}
}
\email{wyx3189967342@bjfu.edu.cn}

\author{Lei Yao}
\affiliation{%
	\institution{University of Wisconsin-Milwaukee}
	\city{Milwaukee}
	\country{USA}
}
\email{leiyao@uwm.edu}

\renewcommand{\shortauthors}{Wang et al.}

\begin{abstract}
Node affinity prediction in the Temporal Graph Benchmark (TGB) asks how a node's future interactions will be distributed over other nodes. On this task, methods that reuse past ground-truth labels often outperform temporal graph neural networks, and the state of the art builds on them. Such comparisons assume a point-in-time correct evaluation, in which a model receives only information available at prediction time. Drawing on as-of joins and bitemporal databases, we formalize this requirement by distinguishing when a label must be predicted, its query time, from when it is fully determined, its availability time, and we audit the four TGB node affinity datasets and their evaluation loop with model-agnostic tools. None of the four meets it. On three datasets, each label summarizes a single day about five days after its timestamp, not the documented following week, yet the loop feeds it back right after that timestamp; on the fourth, the loop processes the edges from which each label is computed before predicting it. Under a drop-in leakage-free protocol, every method whose inputs leak loses up to 17\% of its NDCG@10, the advantage of past labels over the interaction stream nearly vanishes or reverses, and both the best configuration of the state of the art and the rankings change; a blend of training-free long-memory statistics, with weights fit on validation data, matches or exceeds the state of the art under both protocols. Point-in-time correctness should therefore be built into benchmarks, and our tools carry over to other benchmarks whose labels aggregate event streams over time windows.
\end{abstract}

\maketitle

\section{Introduction}
Temporal graph learning supports applications ranging from recommendation and social media analysis to financial and trade forecasting~\cite{kumar2019jodie, trivedi2019dyrep, xu2020tgat, rossi2020tgn, kazemi2020representation}, and dedicated systems such as TGL~\cite{zhou2022tgl} and ETC~\cite{gao2024etc} train its models at scale. Benchmarks guide method development and model selection~\cite{poursafaei2022towards, huang2023temporal, yu2023towards, gastinger2024tgb, yi2025tgbseq, yang2025towards}; among them, the Temporal Graph Benchmark (TGB)~\cite{huang2023temporal} has been widely adopted. Its node affinity prediction task asks how a node's future interactions will be distributed over other nodes, for example, how a user's listening will be spread across music genres~\cite{huang2023temporal, feldman2026revisiting}.

On this task, heuristics that simply reuse past ground-truth labels often outperform temporal graph neural networks (TGNNs)~\cite{huang2023temporal}, and the state of the art, NAVIS~\cite{feldman2026revisiting}, generalizes them as learnable linear state space models; its authors attribute the gap to the inductive biases and training objectives of TGNNs. These explanations take the evaluation protocol for granted, although temporal graph evaluation is prone to temporal leakage, i.e., the use of information that would not be available at prediction time~\cite{kaufman2011leakage, kapoor2023leakage}. Prior reports concern the edges that a model receives: batch-based evaluation distorts link prediction results~\cite{lampert2026link}, a TGB evaluation batch can span multiple days~\cite{tgbissue53}, and on tgbn-trade the reference evaluation loop ingests the target year's edges before predicting that year's labels~\cite{chmura2026tgm, tgmparity}. Whether the labels fed back to models are available when they are used has not been examined, so it remains unclear how much of the effectiveness of past labels reflects genuine temporal regularities rather than artifacts of the evaluation protocol.

\begin{figure}[t]
	\centering
	\includegraphics[width=\columnwidth]{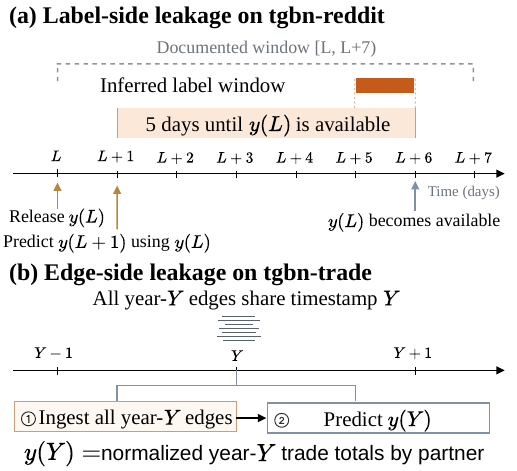}
	\caption{Temporal leakage in TGB: (a) label-side leakage on tgbn-reddit; (b) edge-side leakage on tgbn-trade.}
	\label{fig:examples}
\end{figure}

To answer this question, we audit the four TGB node affinity datasets and their evaluation protocol for point-in-time correctness, which feature stores enforce through as-of joins~\cite{liu2023optimizing}. We associate every label with two times: its \emph{query time}, at which a model must predict it, and its \emph{availability time}, by which all interactions that it summarizes have occurred, that is, at which its valid time in the sense of temporal databases ends~\cite{snodgrass1985taxonomy}. A protocol is leakage-free only if, at each query time, the model has processed no edge from that time onward and has received no label that is not yet available. Violations thus fall into \emph{edge-side leakage}, where edges at or after the query time are processed before prediction, and \emph{label-side leakage}, where labels are fed back before they become available (Figure~\ref{fig:examples}). Label-side leakage does not depend on batching, the focus of prior work~\cite{lampert2026link}, and can arise in any benchmark that feeds window-aggregated labels back to models. Point-in-time correctness is thus a data management problem: it is a property of the pipeline that turns an event stream into model inputs and labels, not of any model, and its violations are invisible at the model level, yet they bias benchmark conclusions. We formalize it with as-of joins and bitemporal time (Section~\ref{sec:pit}), check it with model-agnostic tools (Sections~\ref{sec:toolkit} and~\ref{sec:audit}), and relate its violations to window semantics in stream processing (Section~\ref{sec:lessons-dm}).

We address three research questions. (RQ1) Do the TGB node affinity datasets and their evaluation loop satisfy point-in-time correctness? (RQ2) How does removing leakage change reported performance and the comparison between label-based and edge-based inputs? (RQ3) What temporal signal remains once leakage is removed? For RQ1, a model-agnostic label-window probe recovers from the labels and edges the window that each label actually summarizes, and an instrumented stream replays the reference loop and measures each violation. For RQ2 and RQ3, a drop-in leakage-free protocol processes only edges that precede each query time and releases each label only at its availability time; we re-evaluate both modes of NAVIS and a family of training-free heuristics under both protocols.

\textbf{None of the four datasets is evaluated in a point-in-time correct way.} On tgbn-genre, tgbn-reddit, and tgbn-token, each label summarizes a single day about five days after its timestamp rather than the documented following week, yet it is fed back to models right after its timestamp (Figure~\ref{fig:examples}a), which, to the best of our knowledge, has not been reported before. On tgbn-trade, each yearly label is the distribution of that year's edges, all of which are processed before prediction (Figure~\ref{fig:examples}b).

\textbf{Each type of leakage inflates the methods that can exploit it.} Removing label-side leakage lowers the NDCG@10 of every label-based method by 3\% to 17\%, while methods that use only edges are essentially unaffected. Edge-side leakage on tgbn-trade lets a parameter-free counter score perfectly and lifts the edge-based mode of NAVIS close to it, whereas TGN gains little~\cite{tgmparity}.

\textbf{The advantage of past labels over edges is largely an artifact.} Under the original protocol, label-based inputs outperform edge-based ones on the daily datasets, for both NAVIS and the heuristics. Without leakage, this advantage reverses on tgbn-genre and nearly vanishes on tgbn-reddit and tgbn-token, the best configuration of the state of the art switches from its label-based to its edge-based mode, and the rankings change.

\textbf{What remains is the long-term interaction profile of each node.} Without leakage, the strongest signals average a node's edges or labels over about a week or more; past labels add little beyond the edges they summarize, and timing within a day adds no measurable gain. A blend of such statistics matches or exceeds NAVIS on every daily dataset. NAVIS still outperforms MA on tgbn-genre and tgbn-reddit, though not on tgbn-token, so leakage distorts scores and input comparisons but does not by itself invalidate learned models.

We summarize our contributions as follows:
\begin{enumerate}
	\item A model-agnostic audit showing that none of the four TGB node affinity datasets is evaluated in a point-in-time correct way (Section~\ref{sec:audit}).
	\item A re-evaluation under a leakage-free protocol that changes scores, input comparisons, and rankings, and an analysis of the signal that remains (Sections~\ref{sec:reeval} and~\ref{sec:signal}).
	\item A toolkit for point-in-time audits, comprising a label-window probe, an instrumented evaluation stream, and a drop-in leakage-free protocol, with recommendations for benchmark maintainers and method developers (Sections~\ref{sec:toolkit} and~\ref{sec:lessons}).
\end{enumerate}

\section{Background and Problem Formulation}
\label{sec:background}
This section introduces the task and evaluation loop of TGB, formalizes point-in-time correctness, and classifies its violations.

\subsection{Node Affinity Prediction on TGB}
\label{sec:task}

\textbf{Temporal graphs.} Following prior work~\cite{huang2023temporal, lampert2026link, yang2025towards}, we represent a continuous-time temporal graph as a chronologically ordered stream of edges $\mathcal{E}=\{(s_i,d_i,t_i,w_i)\}_{i=1}^{m}$, where edge $i$ records an interaction from node $s_i$ to node $d_i$ at time $t_i$ with weight $w_i$. We write $\mathcal{E}_{<t}$ for the edges stamped before $t$.

\textbf{Affinity labels.} For a time window $W=[t_s,t_e)$, the affinity vector of a node $u$ assigns to each candidate $v$ the share of the total interaction weight of $u$ in $W$ that goes to $v$~\cite{huang2023temporal}:
\begin{equation}
	\mathbf{y}_u(W)[v] \;=\; \frac{\sum_{(u,v,t_i,w_i)\in\mathcal{E}:\; t_i\in W} w_i}{\sum_{(u,v',t_i,w_i)\in\mathcal{E}:\; t_i\in W} w_i},
	\label{eq:label}
\end{equation}
where $w_i$ is the attribute that the labels aggregate, which may differ from the released edge feature (Section~\ref{sec:audit-window}). TGB stamps each label with a time $t$ and documents it as the affinity vector over the following period of length $k$, one week on the three daily datasets and one year on tgbn-trade~\cite{huang2023temporal}: writing $\mathbf{y}_u(t)$ for the label of $u$ stamped at $t$, $\mathbf{y}_u(t)=\mathbf{y}_u([t,t+k))$, which Section~\ref{sec:audit} examines.

\textbf{Task and metric.} For each label $\mathbf{y}_u(t)$, a model scores all candidates, and its scores $\hat{\mathbf{y}}_u(t)$ are evaluated with NDCG@10~\cite{jarvelin2002cumulated}, which rewards ranking the candidates with the largest true affinities among the top ten. The edges are split chronologically into 70\%, 15\%, and 15\% for training, validation, and testing~\cite{huang2023temporal}.

\textbf{Datasets.} Table~\ref{tab:datasets} summarizes the four datasets: annual agricultural trade flows between nations (tgbn-trade) and users who listen to music genres (tgbn-genre), post in subreddits (tgbn-reddit), or transact cryptocurrency tokens (tgbn-token).

\begin{table}[t]
	\caption{TGB node affinity datasets~\cite{huang2023temporal}; \#Steps: number of unique timestamps.}
	\label{tab:datasets}
	\centering
	\small
	\setlength{\tabcolsep}{3pt}
	\begin{tabular}{@{}lrrrll@{}}
		\toprule
		Dataset & \#Nodes & \#Edges & \#Steps & Interval & Window \\
		\midrule
		tgbn-trade  & 255      & 468{,}245      & 32             & Yearly & 1 year \\
		tgbn-genre  & 1{,}505  & 17{,}858{,}395 & 133{,}758      & Daily  & 1 week \\
		tgbn-reddit & 11{,}766 & 27{,}174{,}118 & 21{,}889{,}537 & Daily  & 1 week \\
		tgbn-token  & 61{,}756 & 72{,}936{,}998 & 2{,}036{,}524  & Daily  & 1 week \\
		\bottomrule
	\end{tabular}
\end{table}

\textbf{Methods by input.} \emph{Label-based} methods consume past ground-truth labels: Persistent Forecast (PF) returns the most recent label of $u$, Moving Average (MA) an exponential moving average of its past labels~\cite{huang2023temporal}, and NAVIS~\cite{feldman2026revisiting} generalizes both as a learnable linear state space model. \emph{Edge-based} methods consume only the edge stream: TGNNs such as TGN~\cite{rossi2020tgn}, DyRep~\cite{trivedi2019dyrep}, and TGNv2~\cite{tjandra2024enhancing} update node representations from incoming edges, and the edge-based mode of NAVIS applies Equation~\eqref{eq:label} to the edges of the previous period~\cite{feldman2026revisiting}. \emph{Hybrid} methods use both.

\subsection{The TGB Evaluation Loop}
\label{sec:loop}

TGB evaluates node affinity methods in a streaming setting~\cite{huang2023temporal}. Algorithm~\ref{alg:loop} summarizes its evaluation loop, simplified from the reference examples.\footnote{\url{https://github.com/shenyangHuang/TGB/tree/main/examples/nodeproppred}} The loop reads the edges in chronological batches of $b$ edges (200 by default) and tracks the current label time $\tau_j$. A batch whose edges are all stamped no later than $\tau_j$ updates the model state (line~\ref{alg:rest}); the first batch with a later edge triggers the prediction (line~\ref{alg:trigger}): the loop passes the edges of this batch that precede the next label time $\tau_{j+1}$ to the model (line~\ref{alg:process}), scores the labels stamped at $\tau_j$ (lines~\ref{alg:predict}--\ref{alg:record}), feeds them back as inputs (line~\ref{alg:observe}), and processes the remaining edges (lines~\ref{alg:remain}--\ref{alg:rest}); with both switches off, as in TGB, no edge is held back and no label is queued. Training uses the same loop with a loss in place of the evaluation; validation and testing may update the model state, but not its parameters~\cite{huang2023temporal}.

\begin{algorithm}[t]
	\caption{TGB evaluation loop for node affinity prediction (simplified). TGB runs it with both switches off; turning them on yields our leakage-free protocol (Section~\ref{sec:protocol}).}
	\label{alg:loop}
	\begin{algorithmic}[1]
		\Require edge batches $B_1, B_2, \dots$ in chronological order; label times $\tau_1 < \tau_2 < \cdots$ and labels $\{\mathbf{y}_u(\tau_j)\}$; model $M$; switches $\mathit{cut}$ and $\mathit{rel}$; availability delay $d$
		\State $j \gets 1$;\ $R \gets \emptyset$;\ $Q \gets \emptyset$ \Comment{held-back edges, label queue}
		\For{each batch $B$}
		\State $A \gets R \cup B$;\ $R \gets \emptyset$\label{alg:merge}
		\If{$\max_{e\in B} t_e > \tau_j$}\label{alg:trigger}
		\State $c \gets \tau_j$ \textbf{if} $\mathit{cut}$ \textbf{else} $\tau_{j+1}$ \Comment{edge cutoff}
		\State $M.\textsc{Process}(\{e\in A : t_e < c\})$\label{alg:process}
		\If{$\mathit{rel}$}
		\While{$Q\ne\emptyset$ \textbf{and} $t_q(\textsc{Front}(Q))+d\le\tau_j$}
		\State $M.\textsc{Observe}(\textsc{PopFront}(Q))$\label{alg:release}
		\EndWhile
		\EndIf
		\For{each node $u$ with a label stamped at $\tau_j$}\label{alg:predict}
		\State $\hat{\mathbf{y}}_u(\tau_j) \gets M.\textsc{Predict}(u)$
		\State record $\mathrm{NDCG@10}\big(\hat{\mathbf{y}}_u(\tau_j), \mathbf{y}_u(\tau_j)\big)$\label{alg:record}
		\EndFor
		\State \textbf{if} $\mathit{rel}$: append $\{\mathbf{y}_u(\tau_j)\}_u$ to $Q$\label{alg:queue}
		\State \textbf{else}: $M.\textsc{Observe}\big(\{\mathbf{y}_u(\tau_j)\}_u\big)$ \Comment{feed labels back}\label{alg:observe}
		\State $A \gets \{e\in A : t_e \ge c\}$;\ $j \gets j+1$\label{alg:remain}
		\EndIf
		\State $c \gets \tau_j$ \textbf{if} $\mathit{cut}$ \textbf{else} $\infty$
		\State $M.\textsc{Process}(\{e\in A : t_e < c\})$;\ $R \gets \{e\in A : t_e \ge c\}$\label{alg:rest}
		\EndFor
	\end{algorithmic}
\end{algorithm}

The loop thus hands models two kinds of future information: it processes edges stamped at or after $\tau_j$ before predicting the labels stamped at $\tau_j$ (lines~\ref{alg:process} and~\ref{alg:rest}), and it feeds each label back right after evaluating it (line~\ref{alg:observe}), although its documented window $[\tau_j,\tau_j+k)$ may extend beyond the next query time $\tau_{j+1}$.

\subsection{Point-in-Time Correctness}
\label{sec:pit}

A forecasting evaluation should give a model only the information that it would have when deployed. We formalize this requirement by attaching two times to every label.

\begin{definition}[Query and availability times]
	\label{def:times}
	Consider a label $\mathbf{y}=\mathbf{y}_u(W)$ with window $W=[t_s,t_e)$, and suppose that a protocol asks a model to predict $\mathbf{y}$ as of time $t_q$. We call $t_q(\mathbf{y})=t_q$ the \emph{query time} of $\mathbf{y}$ and $t_a(\mathbf{y})=t_e$ its \emph{availability time}, i.e., the earliest time by which all interactions summarized by $\mathbf{y}$ have occurred.
\end{definition}

In TGB, labels are stamped with their query time: the loop asks for $\mathbf{y}_u(t)$ as of $t$, although it computes the prediction only after a later edge has arrived. The availability time, in contrast, depends on the window that the label actually summarizes; under the documented window $[t,t+k)$, it is $t+k$.

\begin{definition}[Point-in-time correct evaluation]
	\label{def:pit}
	For a label $\mathbf{y}$, let $\mathcal{I}(\mathbf{y})$ denote the edges and labels that the model has received as inputs before predicting $\mathbf{y}$. An evaluation protocol is \emph{point-in-time correct}, or \emph{leakage-free}, if for every label $\mathbf{y}$
	(i) every edge in $\mathcal{I}(\mathbf{y})$ belongs to $\mathcal{E}_{<t_q(\mathbf{y})}$, and
	(ii) every label $\mathbf{y}'$ in $\mathcal{I}(\mathbf{y})$ satisfies $t_a(\mathbf{y}')\le t_q(\mathbf{y})$.
	A violation of condition (i) is called \emph{edge-side leakage}, and a violation of condition (ii) is called \emph{label-side leakage}.
\end{definition}

Condition (i) is strict because the documented window of a label includes its query time, so an edge stamped exactly at $t_q$ may contribute to the label being predicted; condition (ii) is not, because a label whose window closes at $t_q$ is fully determined by then. Both conditions constrain only what a model receives and when, so they apply to any architecture.

\textbf{Relation to data management.} Definition~\ref{def:pit} is the evaluation-time counterpart of the point-in-time correctness that feature stores guarantee with as-of joins, which attach to each label row only the feature records stamped no later than the label~\cite{liu2023optimizing}. Condition (i) imposes this requirement on the edge stream, with a strict inequality. Condition (ii) covers labels that are themselves fed back as inputs: an as-of join on their stamp would admit labels whose content lies in the future, so they must be joined on their availability time instead. In temporal-database terms~\cite{snodgrass1985taxonomy}, the window of a label is its valid time and the time at which it is fed back is its transaction time, which condition (ii) forbids to precede the end of the valid time. Kaufman et al.\ likewise argue that leakage is best prevented through the way data are managed~\cite{kaufman2011leakage}.

\subsection{A Taxonomy of Temporal Leakage}
\label{sec:taxonomy}

Definition~\ref{def:pit} distinguishes two classes of leakage by the path through which future information reaches the model. Checking Algorithm~\ref{alg:loop} against the two conditions reveals three concrete forms, which Table~\ref{tab:taxonomy} summarizes.

\begin{table}[t]
	\caption{Forms of temporal leakage in the TGB evaluation loop.}
	\label{tab:taxonomy}
	\centering
	\small
	\setlength{\tabcolsep}{3pt}
	\begin{tabular}{@{}>{\raggedright\arraybackslash}p{1.6cm}>{\centering\arraybackslash}p{0.8cm}>{\raggedright\arraybackslash}p{2.8cm}>{\raggedright\arraybackslash}p{1.95cm}@{}}
		\toprule
		Form & Cond. & Mechanism (Alg.~\ref{alg:loop}) & Exploited by \\
		\midrule
		Same-timestamp edges & (i) & Cannot trigger the prediction, so all are processed before it (lines~\ref{alg:process} and~\ref{alg:rest}) & Methods that aggregate recent edges \\
		\addlinespace
		Batch-boundary edges & (i) & Edges stamped after $t_q$ in the triggering batch and in the rest of the previous one are processed first (lines~\ref{alg:process} and~\ref{alg:rest}) & Methods that aggregate recent edges \\
		\addlinespace
		Early label release & (ii) & Labels are fed back right after evaluation, before $t_a$ (line~\ref{alg:observe}) & Methods that consume past labels \\
		\bottomrule
	\end{tabular}
\end{table}

\textbf{Edge-side leakage.} Two kinds of edges stamped at or after $t_q$ can reach the model before the prediction at $t_q$. \emph{Same-timestamp edges} are stamped exactly at $t_q$ and may belong to the window of the label being predicted; when the whole window shares one timestamp, as on tgbn-trade, all of its edges do~\cite{tgmparity}. \emph{Batch-boundary edges} are stamped after $t_q$ and come from the triggering batch and the rest of the previous one, at most $2b$ per query time; they reach beyond one label interval when a batch spans several, because the loop advances by only one label time per batch~\cite{tgbissue53}.

\textbf{Label-side leakage.} \emph{Early label release} feeds a label back before its availability time, which violates condition~(ii) whenever the interval between label stamps is shorter than $t_a - t_q$. Under the documented windows, this holds for the three daily datasets, whose labels are stamped daily but describe a week, but not for tgbn-trade, whose yearly labels become available exactly at the next query time; the accuracy of the window affects only the size of the gap (Section~\ref{sec:audit}).

\textbf{Which methods can exploit leakage.} Label-side leakage enters only through fed-back labels and can thus benefit label-based and hybrid methods but not edge-based ones. Edge-side leakage benefits a method the more directly it turns the leaked edges into an affinity estimate: when they cover the entire label window, summing and normalizing their weights reproduces the label exactly (Equation~\eqref{eq:label}), whereas a memory-based TGNN may gain little~\cite{tgmparity}. Leakage can therefore bias comparisons between input designs and between architectures, not only absolute scores.

\section{Audit Toolkit}
\label{sec:toolkit}
Our toolkit operationalizes Definition~\ref{def:pit} with three model-agnostic components: a label-window probe that recovers the availability time of the labels, an instrumented evaluation stream that measures the violations of the TGB loop, and a leakage-free protocol obtained by two switches of this stream.

\subsection{Label-Window Probe}
\label{sec:probe}

Condition~(ii) depends on the window that each label actually summarizes. This window may deviate from the documentation, and downstream users often receive only the precomputed labels, so the probe recovers it from the labels and the edge stream alone. If a label summarizes a window $W$, it equals the affinity vector that Equation~\eqref{eq:label} computes over $W$, provided that the same edge attribute is aggregated; the probe thus compares each label with the edges of candidate windows at different offsets from its stamp.

\textbf{Similarity and coverage.} Let $\Delta$ be one day for the daily datasets and one year, the label interval, for tgbn-trade. For a label $\mathbf{y}_u(t)$ and an integer offset $a$, let $W_a(t)=[t+a\Delta,\ t+(a+1)\Delta)$, and let $\mathbf{x}^{\omega}_u(W)$ be the vector of total edge weights from $u$ to each candidate in $W$ under a weighting $\omega$, i.e., Equation~\eqref{eq:label} without the normalization. Over a set $\mathcal{T}$ of sampled label times and sets $\mathcal{U}_t$ of sampled labeled nodes, the probe computes the mean similarity
\begin{equation}
	s_{\omega}(a) \;=\; \frac{1}{|\mathcal{T}|}\sum_{t\in\mathcal{T}}\frac{1}{|\mathcal{U}_t|}\sum_{u\in\mathcal{U}_t} \cos\!\big(\mathbf{y}_u(t),\ \mathbf{x}^{\omega}_u(W_a(t))\big),
	\label{eq:probe-sim}
\end{equation}
with the cosine set to zero if $u$ has no edges in $W_a(t)$, and the coverage $c(a)$, the fraction of sampled labeled nodes with at least one edge in $W_a(t)$. The cosine ignores normalization and rescaling of the weights. Coverage adds a sharper test: a label exists only for a node that interacts during its window, since Equation~\eqref{eq:label} is undefined otherwise, so the correct window has a coverage of one. We sample 20 label times from the middle 80\% of the label sequence of each dataset and up to 300 labeled nodes per label time, and scan the offsets $a=-8,\dots,8$.

\textbf{Reading the probe.} Because affinities persist, $s_{\omega}(a)$ is positive at all offsets, so the window shows in the shape of the curves: the documented window $[t,t+k)$ would produce a plateau over the offsets $0,\dots,k/\Delta-1$, whereas a single summarized interval makes the similarity peak and the coverage reach one at the same offset. The probe then takes, among the contiguous windows of at most seven intervals that contain the peak offset and have a coverage of at least 0.99, the one with the highest similarity and denotes by $a^\ast$ its last offset, so that $t_a=t+(a^\ast+1)\Delta$. When a label straddles two probe intervals, for example because of different time zones, this rule yields a conservative availability time. Only the end of the selected window determines $t_a$; because affinities persist, the similarity can favor windows wider than the label, so we do not interpret the start of the window.

\textbf{Edge attributes.} Since a label generator may aggregate another attribute than a method uses, we compute the similarity under unit weights and under the absolute edge feature and use the weighting with the higher peak; if neither reaches one at an aligned peak, the labels aggregate a transformation of the edge feature or another attribute (Section~\ref{sec:audit-window}).

\textbf{Cost and scope.} The probe is linear in the number of probed edges and, since it uses only labels and edges, applies to any benchmark whose labels aggregate the interactions of each entity over a window.

\subsection{Instrumented Evaluation Stream}
\label{sec:instrument}

Whether and how severely a protocol violates Definition~\ref{def:pit} depends on how its loop interleaves edges, predictions, and labels. Our evaluation stream replays the reference examples of TGB and NAVIS exactly, with the same batches, label times, and labeled nodes; models consume it through the three calls of Algorithm~\ref{alg:loop}, \textsc{Process}, \textsc{Predict}, and \textsc{Observe}, so any method that implements them runs on it unchanged.

\textbf{Edge-side diagnostics.} For each query time $t_q$, the stream records the number $n_{\ge}(t_q)$ of edges $e$ with timestamp $t_e \ge t_q$ among the edges $\mathcal{E}_{\mathrm{in}}(t_q)$ ingested before the prediction, and the \emph{edge horizon} $\delta(t_q)$, i.e., how far the latest of them lies beyond $t_q$:
\begin{equation}
	\begin{aligned}
		n_{\ge}(t_q) &= \big|\{e \in \mathcal{E}_{\mathrm{in}}(t_q) : t_e \ge t_q\}\big|,\\
		\delta(t_q) &= \max_{e\in\mathcal{E}_{\mathrm{in}}(t_q)} t_e - t_q.
	\end{aligned}
	\label{eq:edge-diag}
\end{equation}
A negative horizon means that no edge-side leakage occurred, $\delta(t_q)=0$ that only same-timestamp edges leaked, and $\delta(t_q)>0$ that batch-boundary edges leaked and how far into the future they reach (Table~\ref{tab:taxonomy}).

\textbf{Label-side diagnostics.} The stream also logs when each label is released. In the TGB loop, a label is released right after it is scored and thus reaches the next prediction, one label interval $\Delta$ later; at that prediction, its \emph{look-ahead}, i.e., the span of future interactions that the label reveals, is $t_a - t_q - \Delta = a^\ast\Delta$.

\subsection{Leakage-Free Protocol}
\label{sec:protocol}

The leakage-free protocol corrects the TGB loop with a minimal change: it keeps the datasets, splits, batches, label times, labeled nodes, and metric of TGB, and changes only when edges and labels reach the model. Each of its two switches in Algorithm~\ref{alg:loop} enforces one condition of Definition~\ref{def:pit}.

\textbf{Strict edge cutoff.} The first switch, $\mathit{cut}$, enforces condition~(i): before predicting the labels stamped at $\tau_j$, the loop passes to the model only the edges stamped before $\tau_j$ (line~\ref{alg:process}) and holds back the rest until the current label time exceeds their timestamps (lines~\ref{alg:merge} and~\ref{alg:rest}), so no edge is discarded.

\textbf{Delayed label release.} The second switch, $\mathit{rel}$, enforces condition~(ii): the loop queues the scored labels instead of feeding them back (line~\ref{alg:queue}) and, before each prediction at $\tau_j$, releases every queued label $\mathbf{y}$ with $t_q(\mathbf{y})+d \le \tau_j$ (line~\ref{alg:release}). Since all labels of a dataset after the first follow the same window rule, $t_a - t_q$ is a constant, the \emph{availability delay} $d$, which we set to $(a^\ast+1)\Delta$ as recovered by the probe; the first label also satisfies $t_a - t_q \le d$. When $d$ does not exceed the interval between label times, as on tgbn-trade, delayed release coincides with immediate feedback, and only the cutoff takes effect.

\begin{proposition}
	\label{prop:pit}
	If both switches are on and $d \ge t_a(\mathbf{y}) - t_q(\mathbf{y})$ for every label $\mathbf{y}$, Algorithm~\ref{alg:loop} is point-in-time correct and predicts the same labels, for the same nodes, in the same batches as with both switches off. If, in addition, $d = t_a(\mathbf{y}) - t_q(\mathbf{y})$ and edges arrive in timestamp order, every edge and every label reaches the model at the earliest query time permitted by Definition~\ref{def:pit}.
\end{proposition}

\begin{proof}[Proof sketch]
	\textsc{Process} receives only edges stamped before the current label time $\tau_j$, which never decreases, and a label $\mathbf{y}$ is released only if $t_a(\mathbf{y}) \le t_q(\mathbf{y})+d \le \tau_j$, so both conditions hold. The trigger in line~\ref{alg:trigger} does not depend on the switches, so both settings predict at the same label times in the same batches. If $d = t_a - t_q$, a queued label is released at the first $\tau_j \ge t_a(\mathbf{y})$, and with ordered edges, the batch that triggers the prediction at $\tau_j$ has delivered every edge stamped before $\tau_j$.
\end{proof}

\textbf{Design choices.} Simpler fixes change the task or leave leakage in place. Restamping each label at its availability time would let a model compute the label from edges that it has already processed, turning forecasting into aggregation as on tgbn-trade; removing label inputs would exclude methods that are legitimate once labels are released on time; and regenerating the labels over their documented windows would change the targets and still leak (Section~\ref{sec:taxonomy}). Our protocol keeps the targets and the prediction schedule, so scores remain comparable label time by label time.

\textbf{Training and model selection.} We apply the protocol in training, validation, and testing alike: a model trained with immediate feedback learns to rely on labels that a leakage-free test withholds, so a test-only correction would measure this mismatch rather than the effect of leakage, and model selection under the original protocol would favor configurations that exploit leakage.

\textbf{Implementation.} The stream implements both switches at negligible overhead. For NAVIS, we add the same switches to its released training script; since each of its modes consumes one type of input, only the cutoff affects its edge-based mode and only delayed release its label-based mode.

\textbf{Validation.} Our NDCG@10 matches the scikit-learn function used by the TGB evaluator to within $10^{-9}$, and synthetic tests confirm the invariants behind Proposition~\ref{prop:pit} and that the protocol removes or reduces the gain from each injected form of leakage. With both switches off, the patched NAVIS script reproduces the published test NDCG@10 of NAVIS to within 0.002 for its label-based mode on tgbn-trade and tgbn-token and for both modes on tgbn-genre (see Section~\ref{sec:reeval-scores} for its edge-based mode on tgbn-trade); with the strict cutoff, its edge-based mode matches the published scores that we use instead (Section~\ref{sec:reeval-setup}), 0.552 on tgbn-reddit and 0.444 on tgbn-token.

\textbf{Scope.} The protocol controls only which edges and labels reach a model as inputs, not how the model uses the targets that it receives for scoring and training; such misuse must be ruled out in the model itself, as we do for NAVIS.\footnote{In its label-based mode, NAVIS forms a global state from the labels of other nodes at the same query time; we replace it with the mean of the most recently released labels, which changes test NDCG@10 by less than 0.001 on tgbn-trade and tgbn-genre.} The protocol is also only as accurate as its availability times, which is why we resolve ambiguous probe results conservatively.

\section{All Four TGB Node Affinity Benchmarks Leak Future Information}
\label{sec:audit}

This section answers RQ1 by recovering the label windows, locating the leakage in the evaluation loop, and quantifying how much information each form reveals.

\subsection{Audit Setup}
\label{sec:audit-setup}

\textbf{Benchmark version.} We audit the four node affinity datasets of TGB v2.3.0~\cite{huang2023temporal}, the latest release, with their version~1 data files and standard splits, under the reference loop of the TGB examples and NAVIS with batches of 200 edges (Section~\ref{sec:loop}). Neither the label-generation script nor the data files have changed since their release in 2023, so the leakage affects all published results obtained with the official files and loop.

\textbf{Three views of each label.} We compare the \emph{documented window} of the labels (Table~\ref{tab:datasets}), the \emph{implemented window}, computed by the label-generation script of TGB, 
and the \emph{realized window}, recovered from the released data by the probe, and regard a window as established only if the implemented and realized windows agree. For the loop, the reference code shows which violations can occur, and the stream measures how often they occur and how far ahead they reach.

\textbf{Probe and stream settings.} We replay the reference loop over the training, validation, and test splits in this order without resetting its state, as TGB does, and, as a negative control, with the strict cutoff. Table~\ref{tab:audit} reports the test split, on which leaderboard results are computed; our artifact reports all splits. Both tools use only NumPy on a single CPU.

\subsection{Recovered Label Windows}

\label{sec:audit-window}

\begin{figure*}[t]
	\centering
	\includegraphics[width=\textwidth]{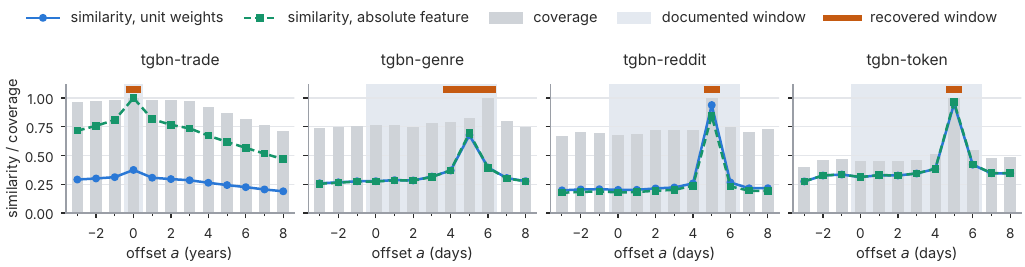}
	\caption{Label-window probe on the four TGB node affinity datasets.}
	\Description{Four panels, one per dataset, plotting similarity and coverage against the offset from the label stamp. The daily datasets show a single sharp peak five days after the stamp, near the end of the shaded documented week instead of a plateau across it; tgbn-trade peaks at offset zero. An orange bar above each panel marks the recovered label window.}
	\label{fig:probe}
\end{figure*}

Figure~\ref{fig:probe} and Table~\ref{tab:audit} show the recovered windows. On the three daily datasets, each label summarizes only one day near the end of its documented week: instead of a plateau over the offsets $0,\dots,6$, both curves remain close to their level at negative offsets and spike only at offset 5 or 6. The window search confirms this: the documented week attains a similarity of 0.59, 0.62, and 0.85 on tgbn-genre, tgbn-reddit, and tgbn-token, against 0.70, 0.94, and 0.97 for the single day five days after the stamp.

\textbf{tgbn-reddit and tgbn-token.} The similarity and the coverage both peak at $a^\ast=5$, where the coverage reaches one, and stay below 0.43 and 0.75 at every other offset: each label stamped at $t$ summarizes the single day $[t+5\Delta,\, t+6\Delta)$ and becomes available at $t_a=t+6\Delta$, as the label-generation script confirms for all labels after the first. 

\textbf{tgbn-genre.} The similarity peaks at $a=5$, whereas the coverage reaches 99.94\% at $a=6$, suggesting that the label day overlaps both probe intervals, consistent with label stamps that are shifted against the day boundaries of the edges, for example by a time-zone conversion; the label-generation script computes these labels like those of tgbn-reddit and tgbn-token.
The interval $[t+5\Delta,\, t+6\Delta)$ attains the highest similarity but covers only 82.7\% of the labeled nodes, and the three best windows with at least 99\% coverage all end at $t+7\Delta$, which gives $a^\ast=6$ and the conservative availability time $t_a=t+7\Delta$.

\textbf{tgbn-trade.} With absolute edge features as weights, the similarity and the coverage reach one at $a^\ast=0$: as documented and as the label-generation script computes it, the label of year $Y$ is the distribution of the trade volume of year $Y$ over partners. 
All of these edges, however, carry the stamp $Y$, which is also the query time of the label.

\textbf{Aggregated attributes.} The two weightings also reveal which attribute the labels aggregate. The absolute edge feature reproduces the labels of tgbn-trade exactly (similarity 1.000) but reaches only 0.965 on tgbn-token, whose labels its exponential reproduces exactly: the loader returns the natural logarithm of the transferred amount that the labels aggregate.\footnote{The released edge list stores the raw amounts; the loader of py-tgb~2.3.0 takes their logarithm only when it reads its cached files, so the first load after a download returns the raw amounts. All our tgbn-token results use the logarithm.} 
On tgbn-reddit, neither weighting reproduces the labels (0.939 and 0.846), so they aggregate an attribute other than the interaction count and the edge feature. 
Edge-based methods thus work with a transformed or approximate version of the label attribute on these two datasets; on tgbn-genre, the misaligned label day prevents this test.

\textbf{Consequences.} On the daily datasets, the benchmark thus asks for the affinity of a single day about five days ahead, whereas the TGB documentation still describes the following week.

\subsection{Leakage in the Evaluation Loop}
\label{sec:audit-loop}

\begin{table}[t]
	\caption{Audit summary on the test split. Forms: S same-timestamp edges, B batch-boundary edges, E early label release; $^\ast$ not reported before, to our knowledge.}
	\label{tab:audit}
	\centering
	\footnotesize
	\setlength{\tabcolsep}{3pt}
	\begin{tabular}{@{}lcccrrrl@{}}
		\toprule
		Dataset & Label window & $t_a-t_q$ & Look-ahead & $\bar n_{=}$ & $\bar n_{>}$ & max $\delta$ & Forms \\
		\midrule
		trade  & $[t,\,t{+}1)$ yr              & 1 yr & none   & 19{,}126 & 0     & 0       & S \\
		genre  & $[t{+}5,\,t{+}7)$ d$^\dagger$ & 7 d  & 5--6 d & 1.6      & 99.8  & 2.3 h   & E$^\ast$, B, S \\
		reddit & $[t{+}5,\,t{+}6)$ d           & 6 d  & 5 d    & 0.7      & 90.9  & 6.1 min & E$^\ast$, B, S \\
		token  & $[t{+}5,\,t{+}6)$ d           & 6 d  & 5 d    & 3.5      & 108.2 & 2.8 min & E$^\ast$, B, S \\
		\bottomrule
	\end{tabular}
	\\[2pt]
	{\footnotesize $^\dagger$A single day within this interval, not aligned with the day grid; the three best windows with at least 99\% coverage all end at $t{+}7$\,d.}
\end{table}

Table~\ref{tab:audit} combines the recovered windows with the diagnostics of the stream, splitting $n_{\ge}$ (Equation~\eqref{eq:edge-diag}) into the edges stamped exactly at the query time, $n_{=}$, and those stamped after it, $n_{>}$. With the strict cutoff, no such edge reaches the model, as expected; the reference loop exhibits all three forms of leakage, with very different severity.

\textbf{Early label release.} On the daily datasets, the loop feeds every label back right after scoring it, so the label reaches the model one day after its stamp, although it becomes available only six to seven days after it: a look-ahead of five days on tgbn-reddit and tgbn-token and five to six days on tgbn-genre. Since each label stays premature for $a^\ast$ query times, a label-consuming model holds several premature labels at once: at a query time $t$ on tgbn-reddit, the labels stamped at $t-5\Delta,\dots,t-\Delta$ give it the affinity of a node for each of the five days $[t,\,t+\Delta),\dots,[t+4\Delta,\,t+5\Delta)$ before the target day, as far as the node interacts on them. On tgbn-trade, a label becomes available exactly at the next query time and thus does not leak.

\textbf{Same-timestamp edges.} On tgbn-trade, all edges of year $Y$ carry the stamp $Y$, the query time of the labels of that year. Because only a batch with a later edge triggers the prediction, the loop processes every edge of year $Y$, 19{,}126 per query time on average on the test split, before it predicts the labels computed from exactly these edges ($\delta=0$), as noted before~\cite{tgmparity}. On the daily datasets, whose timestamps are given in seconds, fewer than four edges per query time carry the query time as their stamp on average.

\textbf{Batch-boundary edges.} On the test splits of the daily datasets, almost all edges that reach the model at or after the query time are batch-boundary edges: 91 to 108 per query time on average, less than 1\% of a day's edges, reaching at most 2.3 hours ahead. When a batch spans several label intervals, as on the training split of tgbn-token, they reach up to 2.2 days with up to $2b=400$ edges, the known issue that a TGB batch can span multiple days~\cite{tgbissue53}. On tgbn-trade, $\bar n_{>}=0$.

\subsection{Information Revealed by Each Form}
\label{sec:audit-reveal}

The diagnostics show how much information leaks, but not how useful it is. To measure this without confounding it with model capacity, we evaluate, for each form, a parameter-free method that reads only the input through which the form leaks, in the spirit of EdgeBank~\cite{poursafaei2022towards}, under the TGB loop with and without this form.

\textbf{Same-timestamp edges.} The edge counter scores each candidate by the total weight of the node's edges to it since the node's previous prediction. On tgbn-trade, these are exactly the edges from which the label is computed, so the counter reproduces every label and attains an NDCG@10 of 1.000 without learning anything; with the strict cutoff, it sees only the previous year and coincides with PF (0.855). Any method that aggregates the edges of the current period is thus rewarded there for aggregation rather than forecasting.

\textbf{Early label release.} PF predicts the most recent label that it has received. Under the TGB loop, it reproduces the values reported in prior work~\cite{feldman2026revisiting}, 0.357 on tgbn-genre and 0.369 on tgbn-reddit; with delayed release, it loses 0.046, 0.061, and 0.068 on tgbn-genre, tgbn-reddit, and tgbn-token, or 13\% to 17\%. Early release turns PF into a one-day-ahead forecast, and most of the inflation stems from the labels of the two days before the target day: releasing labels three days after their stamp, which withholds exactly these two, already removes about 90\% of it on tgbn-reddit (Section~\ref{sec:reeval-robust}).

\textbf{Batch-boundary edges.} The strict cutoff changes the edge counter and a recency counter, which decays each edge weight exponentially with a time constant of 3 hours, by at most 0.0002 on all three daily datasets: these edges occur within hours after the query time, whereas the label summarizes a day about five days later. Batch-boundary leakage is thus negligible here, unlike in link prediction, where batch-based evaluation distorts results~\cite{lampert2026link}.

\textbf{Answer to RQ1.} None of the four datasets is evaluated in a point-in-time correct way. Two forms of leakage matter: on the daily datasets, early label release turns label-based forecasting into a one-day-ahead forecast, and on tgbn-trade, same-timestamp edges make the task trivial for any method that aggregates the edges of the current period. Batch-boundary edges are negligible.

\section{Leakage Distorts the TGB Node Affinity Leaderboard}
\label{sec:reeval}

This section answers RQ2: how removing leakage changes scores, the comparison between label-based and edge-based inputs, and rankings.

\subsection{Experimental Setup}
\label{sec:reeval-setup}

\textbf{Methods.} The \emph{label-based} methods are the TGB heuristics PF and MA, EMA ($\beta=0.95$), an exponential moving average with longer memory, a blend of label channels (the \emph{label blend}), and the label-based mode of NAVIS~\cite{feldman2026revisiting}, the state of the art. As released in TGB, MA is an exponential moving average, $h_u \gets \beta\,h_u + (1-\beta)\,y$ over the labels $y$ of $u$ starting from the first one, with $\beta=(k-1)/k$ and $k=6$ on tgbn-genre and $k=7$ otherwise.\footnote{The TGB paper describes MA as the average of the previous $k=7$ labels~\cite{huang2023temporal}; with the released code, we reproduce the leaderboard values of PF and MA on tgbn-genre and tgbn-trade.} 
The label channels are PF and the moving averages with $\beta\in\{0.5, 0.7, (k-1)/k, 0.95\}$.
The edge channels aggregate the edge weights between a node and a candidate since the node's previous prediction (the edge counter), over the whole history (the total edge weight), as moving averages of per-period totals ($\beta\in\{0.5, 6/7\}$), and with exponential decay at time constants of $1/8$ (the recency counter), $1/2$, $2$, and $8$ label intervals. The \emph{hybrid blend} combines both. Each blend is a non-negative linear combination of row-normalized channels, fit by coordinate ascent on NDCG@10 over up to 60 validation label times with 400 nodes each, separately under each protocol. Edges are weighted by their absolute feature, except for the heuristics on tgbn-reddit, where unit weights approximate the labels more closely (Section~\ref{sec:audit-window}).

\textbf{Protocols.} The \emph{original} protocol runs Algorithm~\ref{alg:loop} with both switches off, the \emph{leakage-free} protocol with both on and the availability delays of Table~\ref{tab:audit}, in training, validation, and testing alike. For NAVIS, only the switch that matches the input of each mode takes effect, and in leakage-free label-based runs we also correct its global state (Section~\ref{sec:protocol}). For the heuristics on the daily datasets, the original protocol also applies the strict cutoff, which changes the blends by at most about 0.001, so that the protocols differ only in label release.

\textbf{Training.} We train NAVIS with its released code and hyperparameters for up to 50 epochs (with early stopping after 10 epochs without improvement on tgbn-token and tgbn-reddit) and report, as its code does, the test score at the epoch with the best validation score, averaged over seeds 1, 2, and 3.
On tgbn-reddit, and for the edge-based mode on tgbn-token, its scores under the original protocol are taken from its paper~\cite{feldman2026revisiting}; our code reproduces all other published scores except that of the edge-based mode on tgbn-trade (Section~\ref{sec:reeval-scores}). 
The heuristics have no trainable parameters apart from the blend weights and are deterministic given the validation sample.

\textbf{Evaluation.} We report test NDCG@10 averaged over label times, as TGB does, and its relative change between the protocols. For the heuristics, we pair two methods or protocols by test label time and report the mean difference with a 95\% bootstrap confidence interval (2{,}000 resamples) and a Wilcoxon signed-rank test, except on tgbn-trade with its 3 test label times. We compare rankings by Kendall's $\tau_b$.

\subsection{Changes in Scores}
\label{sec:reeval-scores}

\begin{table*}[t]
	\caption{Test NDCG@10 under the original (Orig.) and leakage-free (LF) protocols; $^\dagger$ reported in~\cite{feldman2026revisiting}. Other NAVIS scores average three seeds.}
	\label{tab:reeval}
	\centering
	\small
	\setlength{\tabcolsep}{3.6pt}
	\begin{tabular}{@{}l rrr rrr rrr rrr@{}}
		\toprule
		& \multicolumn{3}{c}{tgbn-trade} & \multicolumn{3}{c}{tgbn-genre} & \multicolumn{3}{c}{tgbn-reddit} & \multicolumn{3}{c@{}}{tgbn-token} \\
		\cmidrule(lr){2-4} \cmidrule(lr){5-7} \cmidrule(lr){8-10} \cmidrule(l){11-13}
		Method & Orig. & LF & $\Delta$ & Orig. & LF & $\Delta$ & Orig. & LF & $\Delta$ & Orig. & LF & $\Delta$ \\
		\midrule
		\multicolumn{13}{@{}l}{\emph{Label-based}} \\
		PF                  & 0.855 & 0.855 & 0.0         & 0.357 & 0.311 & $-$12.9 & 0.369 & 0.308 & $-$16.5 & 0.430 & 0.362 & $-$15.7 \\
		MA                  & 0.823 & 0.823 & 0.0         & 0.509 & 0.476 & $-$6.5  & 0.559 & 0.531 & $-$5.0  & 0.508 & 0.451 & $-$11.3 \\
		EMA ($\beta=0.95$)  & 0.725 & 0.725 & 0.0         & 0.530 & 0.511 & $-$3.6  & 0.572 & 0.555 & $-$2.9  & 0.486 & 0.440 & $-$9.4 \\
		Label blend         & 0.858 & 0.858 & 0.0         & 0.532 & 0.511 & $-$4.0  & 0.576 & 0.555 & $-$3.7  & 0.513 & 0.451 & $-$12.0 \\
		NAVIS (label mode)  & 0.863 & 0.863 & 0.0         & 0.527 & 0.501 & $-$5.0  & 0.569\rlap{$^\dagger$} & 0.549 & $-$3.5 & 0.513 & 0.439 & $-$14.4 \\
		\midrule
		\multicolumn{13}{@{}l}{\emph{Edge-based}} \\
		NAVIS (edge mode)   & 0.998 & 0.863 & $-$13.5     & 0.520 & 0.520 & 0.0     & 0.552\rlap{$^\dagger$} & 0.552 & 0.0     & 0.444\rlap{$^\dagger$} & 0.444 & 0.0 \\
		Edge blend          & 1.000 & 0.863 & $-$13.7     & 0.527 & 0.527 & 0.0     & 0.554 & 0.554 & 0.0     & 0.447 & 0.447 & 0.0 \\

		\midrule
		\multicolumn{13}{@{}l}{\emph{Hybrid}} \\
		Hybrid blend        & 1.000 & 0.863 & $-$13.7     & 0.542 & 0.530 & $-$2.2  & 0.580 & 0.561 & $-$3.3  & 0.513 & 0.454 & $-$11.5 \\
		\bottomrule
	\end{tabular}
\end{table*}

\textbf{Label-based methods decline by 3\% to 17\% on the daily datasets.} PF declines most, by 13\% to 17\% (Table~\ref{tab:reeval}), because it relies on the most recent label alone, which early release supplies five to six days early (Section~\ref{sec:audit-reveal}). Methods that also draw on older labels decline by 2.9\% to 6.5\% on tgbn-genre and tgbn-reddit, but by 9.4\% to 14.5\% on tgbn-token, the only daily dataset on which MA, with its shorter memory, outperforms EMA ($\beta=0.95$) under the original protocol, which indicates that most of the signal there lies in the most recent labels. For every label-based heuristic on every daily dataset, the 95\% confidence interval of the paired difference excludes zero (Wilcoxon $p<10^{-6}$). 

\textbf{Edge-based methods are unaffected on the daily datasets.} The edge blend sees only edges stamped before the query time under both protocols, and the edge-based mode of NAVIS, which under the original protocol also receives the batch-boundary edges, has the same score to three decimal places under both protocols. The hybrid blend declines by 2.2\% to 11.5\%, less than the label blend on every daily dataset, because its fitted weights shift toward the unchanged edge channels; on tgbn-genre, for example, the total weight of its label channels falls from 0.53 to 0.32. 

\textbf{On tgbn-trade, near-perfect scores fall to the level of PF.} The label-based methods are unaffected here, because the TGB loop already uses each yearly label only from the next label time on, whereas every method that aggregates the edges of the current year declines by 13.5\% to 13.7\%, from 1.000 or 0.998 to 0.863, close to PF (0.855). The published value for the edge-based mode of NAVIS, 0.863~\cite{feldman2026revisiting}, equals both its published label-based score and our leakage-free value, rather than the 0.998 that the released code attains under the reference loop; we could not determine which configuration produced it.

\subsection{Label-Based versus Edge-Based Inputs}
\label{sec:reeval-inputs}

\begin{figure}[t]
	\centering
	\includegraphics[width=\columnwidth]{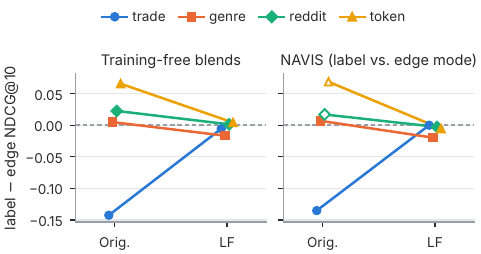}
	\caption{Advantage of past labels over edges under the original and leakage-free protocols.}
	\label{fig:inputs}
\end{figure}

Whether past labels or the interaction stream are more informative is a recurring question: TGB~\cite{huang2023temporal} and NAVIS~\cite{feldman2026revisiting} report that heuristics built on past labels outperform TGNNs, and NAVIS scores higher in its label-based than in its edge-based mode on every daily dataset. 
Since comparing heuristics with TGNNs confounds the input with the model, Figure~\ref{fig:inputs} compares the inputs with the model held fixed, for the blends and for the two modes of NAVIS.

\textbf{Under the original protocol, the input that leaks wins.} On the daily datasets, where labels leak, label-based methods lead: the label blend by 0.005, 0.023, and 0.066 on tgbn-genre, tgbn-reddit, and tgbn-token, and the label-based mode of NAVIS by 0.007, 0.017, and 0.069. On tgbn-trade, where edges leak, edge-based methods lead by 0.14 in both settings.

\textbf{Without leakage, the advantage of past labels nearly vanishes or reverses.} The two inputs then lie within 0.02 of each other on every dataset. On tgbn-genre, edge-based methods now lead, by 0.017 (95\% CI $[0.014, 0.019]$) for the blends and by 0.020 for NAVIS. On tgbn-reddit and tgbn-token, the advantage of the label blend shrinks by 94\% and 93\%, to 0.001 and 0.004, which remain significant but are an order of magnitude smaller, and the edge-based mode of NAVIS now leads by 0.003 and 0.005. On tgbn-trade, the inputs tie to within 0.006, as expected: without same-timestamp edges, the edges of the previous year reproduce its label exactly (Section~\ref{sec:audit-reveal}).

\textbf{Both settings agree.} The blends and NAVIS share neither parameters nor training procedure, yet show the same pattern, and since the two modes of NAVIS differ only in their input, the original advantage of past labels stems from leakage, not from the model. Comparisons between label-based and edge-based methods, including those between the heuristics and the TGNNs on the leaderboard, thus require a point-in-time correct protocol; on tgbn-reddit, the remaining difference also depends on the edge weighting (Section~\ref{sec:reeval-robust}).

\subsection{Changes in Rankings}
\label{sec:reeval-ranks}

\begin{table}[t]
	\caption{Ranks under the original and leakage-free protocols (Orig.~$\to$~LF); ties at three decimals share a rank.}
	\label{tab:ranks}
	\centering
	\small
	\setlength{\tabcolsep}{5pt}
	\begin{tabular}{@{}lcccc@{}}
		\toprule
		Method & trade & genre & reddit & token \\
		\midrule
		PF                 & 6 $\to$ 6 & 8 $\to$ 8 & 8 $\to$ 8 & 8 $\to$ 8 \\
		MA                 & 7 $\to$ 7 & 7 $\to$ 7 & 5 $\to$ 7 & 4 $\to$ 2 \\
		EMA ($\beta=0.95$) & 8 $\to$ 8 & 3 $\to$ 4 & 3 $\to$ 2 & 5 $\to$ 6 \\
		Label blend        & 5 $\to$ 5 & 2 $\to$ 4 & 2 $\to$ 2 & 1 $\to$ 2 \\
		NAVIS (label mode) & 4 $\to$ 1 & 4 $\to$ 6 & 4 $\to$ 6 & 1 $\to$ 7 \\
		\midrule
		NAVIS (edge mode)  & 3 $\to$ 1 & 6 $\to$ 3 & 7 $\to$ 5 & 7 $\to$ 5 \\
		Edge blend         & 1 $\to$ 1 & 4 $\to$ 2 & 6 $\to$ 4 & 6 $\to$ 4 \\
		\midrule
		Hybrid blend       & 1 $\to$ 1 & 1 $\to$ 1 & 1 $\to$ 1 & 1 $\to$ 1 \\
		\midrule
		Kendall's $\tau$   & 0.90 & 0.59 & 0.69 & 0.46 \\
		\bottomrule
	\end{tabular}
\end{table}

\textbf{Rankings change on every daily dataset, along the input of each method.} Table~\ref{tab:ranks} ranks the methods under both protocols; Kendall's $\tau$ between the two rankings is 0.59, 0.69, and 0.46 on tgbn-genre, tgbn-reddit, and tgbn-token. Both edge-based methods rise by two to three ranks, whereas the label-based mode of NAVIS falls, from fourth to sixth on tgbn-genre and tgbn-reddit and from first to seventh on tgbn-token. On tgbn-trade, $\tau$ is 0.90, but the edge aggregators that lead with near-perfect scores under the original protocol tie with the label-based mode of NAVIS without leakage.

\textbf{The best configuration of the state of the art changes.} Under the original protocol, NAVIS performs best with past labels on every daily dataset, as its paper reports~\cite{feldman2026revisiting}; without leakage, its edge-based mode is better on all three, by 0.020, 0.003, and 0.005 (0.008 on tgbn-reddit with unit edge weights). A practitioner who follows the leaderboard would thus choose the configuration that performs worse under point-in-time correct evaluation. 

\textbf{Training-free blends match or exceed the state of the art.} Without leakage, the hybrid blend ranks first on every daily dataset, ahead of the better mode of NAVIS by 0.010 on tgbn-genre and tgbn-token and by 0.004 on tgbn-reddit, where we compare against the edge-based mode with unit weights. It already leads by 0.015 and 0.011 under the original protocol on tgbn-genre and tgbn-reddit, whereas on tgbn-token, where the two tie, the lead emerges because NAVIS loses more without leakage. 

The changes do not depend on our blends, which are not on the TGB leaderboard: among PF, MA, and the two modes of NAVIS alone, the first place changes on every daily dataset, from the label-based mode of NAVIS to its edge-based mode on tgbn-genre and tgbn-reddit and to MA on tgbn-token.

\subsection{Robustness of the Findings}
\label{sec:reeval-robust}

\textbf{Availability delay.} The protocol releases each label at the end of its recovered window, on tgbn-genre at the end of the later calendar day. The results hardly depend on the exact delay: on tgbn-reddit, releasing labels after three days, half the availability delay, already produces 70\% to 90\% of the full decline of each method that reads labels, and the remaining three days lower each score by at most 0.007. On tgbn-genre, the label blend trails the edge blend significantly at delays of five, six, and seven days alike, so the reversal there does not depend on which calendar day ends the label window. 

\textbf{Edge weighting.} On tgbn-reddit, the blends use unit edge weights and NAVIS absolute ones. Unit weights help both edge-based methods, the edge blend by 0.016 and NAVIS by 0.005, in line with the probe, so they suit the dataset, not just our blends. Of the findings above, they affect only how much the advantage of past labels shrinks: with absolute weights, it falls from 0.038 to 0.017, by more than half rather than by 94\%.

\textbf{Seeds.} For the NAVIS configurations in Table~\ref{tab:reeval} evaluated over three seeds, the standard deviation is at most 0.00041, against declines of 0.020 to 0.074 for its label-based mode on the daily datasets. Without leakage, its edge-based mode leads on tgbn-reddit and tgbn-token for all three seeds (by 0.003 and 0.005 on average).

\textbf{Epoch selection.} We report NAVIS at the epoch with the best validation score under both protocols, as its released code does; at epoch 50 for seed 2, its leakage-free edge-based scores on tgbn-reddit would be about 0.025 lower, which would only widen the lead of the training-free blends.

\textbf{Answer to RQ2.} Leakage inflates every method whose inputs carry leaked information about the target: removing it lowers label-based methods on the daily datasets by 3\% to 17\% and edge aggregation on tgbn-trade by about 14\%, and leaves the other methods essentially unchanged. Because it favors a different input on each kind of dataset, it largely determines the comparison between the inputs: without leakage, the advantage of past labels over edges nearly vanishes or reverses, the best configuration of the state of the art switches to its edge-based mode, and the rankings change ($\tau$ between 0.46 and 0.69 on the daily datasets). These findings are robust to the availability delays and to the seeds of NAVIS; the edge weighting changes only how much the advantage of past labels shrinks on tgbn-reddit.

\section{Without Leakage, Long-Term Interaction Statistics Carry the Signal}
\label{sec:signal}

This section answers RQ3 on the three daily datasets with the training-free channels of Section~\ref{sec:reeval-setup}, without leakage; they do not exhaust what a learned model could extract, but show how far back the useful information reaches, whether past labels add to the edges, and whether timing matters.

\subsection{Long Memory Matters More than Recency}
\label{sec:signal-memory}

\begin{figure}[t]
	\centering
	\includegraphics[width=\columnwidth]{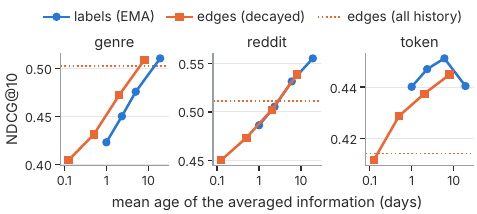}
	\caption{Test NDCG@10 of single channels by memory length without leakage.}
	\label{fig:memory}
\end{figure}

Figure~\ref{fig:memory} places each channel at the mean age of the information it averages: $\beta/(1-\beta)$ label intervals for a moving average of labels with factor $\beta$, counted from the newest released label, whose window ends at about the query time, and the time constant for a decayed edge weight.

\textbf{Longer memory scores higher.} The decayed edge weights improve monotonically from a time constant of 3 hours to 8 days on all three datasets, and the moving averages of labels up to a mean age of 19 days on tgbn-genre and tgbn-reddit, whereas PF and the edge counter, which rely on the most recent information alone, score lowest. Memory has a limit, however: the total edge weight over the whole history scores below the edge weights decayed over 8 days on all three datasets.

\textbf{The blends rely almost entirely on long memory.} In the leakage-free hybrid blend, channels with a mean age of at least six days (the label moving averages with $\beta\geq 6/7$, the total edge weight, the moving average of per-period edge totals with $\beta=6/7$, and the edge weights decayed over 8 days) receive 96\% of the weight on tgbn-genre and all of it on tgbn-reddit and tgbn-token.

\textbf{The forecast horizon likely explains this preference.} Without leakage, the label window begins five to six days after the newest usable information (Section~\ref{sec:audit-window}), a horizon over which short-term fluctuations rarely persist, whereas long-term preferences do. Leakage shortens this horizon and thus favors short memory: removing it lowers PF by 13\% to 17\%, but EMA ($\beta=0.95$) by only 3\% to 9\%.

\textbf{Preferences appear to change faster on tgbn-token.} There, both curves are flatter and the label moving average peaks at a mean age of six days, consistent with the larger declines of the label-based methods; yet the best channels still average over several days.

\subsection{Past Labels Add Little Beyond the Edges}
\label{sec:signal-labels}

In these benchmarks, the labels aggregate the same interactions that form the edge stream. Once released at its availability time, a label summarizes edges that an edge-based method has already received, so past labels can add information beyond the edges only through how they aggregate them: which attribute they sum, how they normalize it, and over which window. Three observations show that this adds little.

\textbf{With matched aggregations, labels beat edges only on tgbn-token.} The moving averages of labels and of per-period edge totals average one normalized vector per period with the same factor $\beta$, so comparing them isolates the input, up to slightly different periods. With $\beta=0.5$, the edge version scores 0.458 against 0.423 on tgbn-genre and 0.494 against 0.486 on tgbn-reddit, and with $\beta=6/7$, both score 0.531 on tgbn-reddit; only on tgbn-token is the label version ahead, by about 0.02 for both factors.

\textbf{Combining both inputs gains less than leakage did.} Adding the label channels to the edge blend raises its score by 0.003 to 0.008, less than half of the 0.021 to 0.062 that leakage added to the label blend, and NAVIS loses nothing when it replaces the labels with the semi-labels that it aggregates from the edges: without leakage, its edge-based mode scores higher on all three datasets (Section~\ref{sec:reeval-ranks}). 

\textbf{Where labels help, they aggregate differently.} The label blend keeps a small advantage over the edge blend only on tgbn-token and tgbn-reddit, whose labels aggregate the transferred amounts, of which the edge feature carries the logarithm, and an attribute other than the edge feature, respectively (Section~\ref{sec:audit-window}). This residual advantage thus most likely stems from attributes that the edge features omit or transform.

\subsection{Recency Across Periods, Not Timing Within Them}
\label{sec:signal-timing}

The exact timestamps of edges, which continuous-time TGNNs are designed to exploit, tell not only on which day an edge occurred but also when within that day; among our channels, only the decayed edge weights see this timing.

\textbf{Timing within a day adds no measurable gain.} Without leakage, the decayed edge weights improve the hybrid blend by 0.0023, 0.0010, and 0.0013 (95\% confidence intervals excluding zero), about a tenth or less of the decline that leakage causes for the label blend. On a placebo stream that permutes the endpoints and weights of the edges over the fixed timestamps within each label period, which preserves every per-period total but destroys the timing within it, they improve the blend almost as much, by 0.0017, 0.0011, and 0.0012, and none of the differences from the real stream is significant (Wilcoxon $p\ge 0.19$). The decayed channels thus help by weighting recent days more heavily than older ones, not by resolving the order of events within a day.

\textbf{Implications for models of the edge stream.} A learned model may extract timing patterns that our channels miss, but at least for the statistics we consider, these benchmarks reward little of what distinguishes TGNNs from simpler aggregators; evaluating that distinction would require targets that depend on fine-grained timing. 

\textbf{Answer to RQ3.} Without leakage, the signal that our channels capture on the daily TGB node affinity datasets is the long-term interaction profile of each node: averages over about a week or more of its edges or labels, with a mild preference for recent days. Past labels add little beyond the edges they summarize, and what they add most likely stems from attributes that the edge features omit or transform. Timing within a day adds no measurable gain beyond the daily totals.

\section{Point-in-Time Correctness Belongs in the Benchmark}
\label{sec:lessons}

\begin{table}[t]
	\caption{Checklist for point-in-time correct evaluation of node affinity prediction. B: benchmark maintainers; D: method developers.}
	\label{tab:checklist}
	\centering
	\footnotesize
	\setlength{\tabcolsep}{3pt}
	\renewcommand{\arraystretch}{1.15}
	\begin{tabular}{@{}>{\raggedright\arraybackslash}p{0.35\columnwidth}c>{\raggedright\arraybackslash}p{0.26\columnwidth}>{\raggedright\arraybackslash}p{0.245\columnwidth}@{}}
		\toprule
		Check & Who & Violated in TGB & How to check \\
		\midrule
		Each label states its window and availability time & B & all; documented window wrong for genre, reddit, token & label-window probe (Section~\ref{sec:probe}) \\
		Labels become inputs only at their availability time & B & genre, reddit, token & delayed release (Section~\ref{sec:protocol}) \\
		Only edges stamped before $t_q$ are processed before predicting at $t_q$ & B & trade (same timestamp); genre, reddit, token (batch boundary) & strict cutoff (Section~\ref{sec:protocol}); loop diagnostics (Section~\ref{sec:instrument}) \\
		Loading returns the same data with and without a cache & B & token (py-tgb 2.3.0) & checksums of the processed data \\
		No model component reads labels before their availability time & D & NAVIS global state (small effect) & ablation; code review \\
		Training, validation, and testing use the same protocol & B, D & -- & -- \\
		Inputs, edge weighting, and epoch selection are reported & D & -- & -- \\
		\bottomrule
	\end{tabular}
\end{table}

With one minor exception in NAVIS, the problems we found lie not in particular models but in how the benchmark records and releases time: labels whose window and availability time are not part of the data, and a loop that hands models their inputs before these would be available. Table~\ref{tab:checklist} condenses our recommendations.

\subsection{Recommendations}
\label{sec:lessons-recs}

\textbf{For benchmark maintainers.} Each label should be released with its window, its valid time in a bitemporal record, and its availability time, the earliest transaction time at which it may be fed back (Section~\ref{sec:pit}), and specified declaratively by its aggregated attribute, normalization, and window relative to the timestamp, which documents the window by construction, as a window clause does for a windowed aggregate query; procedural generation code let the documentation and the data of TGB diverge unnoticed. The two switches of Algorithm~\ref{alg:loop} keep the schedule of the reference loop and can thus become its default. TGM already applies the strict cutoff by default~\cite{chmura2026tgm, tgmparity}, so delayed release is the missing half, and on the daily datasets the half that matters; frameworks that load TGB inherit its label windows and thus its availability delays. 
Running the probe and the loop diagnostics at each release, for instance in continuous integration, would have revealed both problems, and applying every loader transformation in one place, with checksums, would have prevented the cache-dependent loading of tgbn-token. The corrected protocol and label metadata should form a new benchmark version, with baselines re-evaluated and earlier results marked as obtained under the original protocol. 

\textbf{For method developers.} Since early labels inflate label-based methods and same-timestamp edges edge-based ones (Section~\ref{sec:reeval-inputs}), papers should state the inputs of each method and when each label is used, and train, select, and test under the same point-in-time correct protocol; statistics computed across nodes, such as the global state of NAVIS, must respect the same availability times. Claims about which input is more informative should rest on comparisons that change only the input (Section~\ref{sec:signal-labels}). The edge weighting and the epoch selection rule, which change our results on tgbn-reddit by 0.016 and 0.025, should be reported with every result, together with early stopping and the number of seeds. Finally, as EdgeBank did for link prediction~\cite{poursafaei2022towards}, training-free long-memory blends expose what the benchmark rewards and should serve as baselines (Section~\ref{sec:reeval-ranks}), and new methods should draw on about a week or more of history (Section~\ref{sec:signal-memory}).

\subsection{Beyond TGB: A Data Management View}
\label{sec:lessons-dm}

Beyond as-of joins and bitemporal data (Section~\ref{sec:pit}), each form of leakage corresponds to a well-understood problem in stream processing, and point-in-time correctness can be checked like other data constraints.

\textbf{Early label release emits window results before their windows close.} Stream processors emit the result of a window aggregate once the watermark, an estimate of how far event time has progressed, passes the end of the window~\cite{akidau2015dataflow}. Each TGB label is such a result, but the loop releases it at its stamp, the start of its documented window, and thus on the daily datasets before its realized window has even begun; delayed release acts as a watermark-driven trigger and, like one, needs to know where each window ends.

\textbf{Edge-side leakage confuses processing order with event time.} The loop triggers predictions by batches, units of processing, and breaks ties between edges and queries with the same timestamp in favor of the edges. Just as stream processors assign events to windows by their timestamps rather than by their arrival, the strict cutoff admits an edge to a query only if its event time precedes the query time, regardless of its batch.

\textbf{Point-in-time correctness can be checked like other data constraints.} Data validation systems check declared constraints on the data that feed ML pipelines~\cite{schelter2018automating, breck2019data}, and leakage detectors analyze pipeline code~\cite{yang2022data}. Point-in-time constraints relate the timestamps of the inputs to those of the queries that consume them and thus concern the evaluation loop as well as the data, but they are nevertheless cheap to check: the loop diagnostics need no model, and the probe needs no training.

\textbf{The pattern recurs wherever labels summarize future events.} Labels such as the purchases of a user or the demand for a product in the coming days, stamped with the time at which they are predicted, become available only after their windows close, so feeding them back at their stamps leaks, and breaking ties in favor of the inputs or processing them in batches can leak the very events from which a label is computed. Our protocol applies to any stream of timestamped events and labels once the availability delays are known, and the probe recovers these delays whenever each label aggregates the interactions of an entity with others.

\subsection{Limitations}
\label{sec:limitations}

\textbf{Scope.} We do not audit the link prediction datasets of TGB and TGB~2.0~\cite{gastinger2024tgb}, whose targets are edges available at their own timestamps, so early label release cannot occur there; batching effects there have been studied by Lampert et al.~\cite{lampert2026link}. Our results refer to TGB v2.3.0 with its version~1 data files and to NAVIS at commit 2b503ef; 
a future release would call for a new audit, which our tools make cheap.

\textbf{Recovered windows depend on the edge stream.} The probe resolves windows only to one day and only if the labels aggregate interactions in the released edge stream, so we regard a window as established only where the label-generation script agrees with it. On tgbn-genre, whose label day straddles two calendar days, we release each label at the end of the later day, the conservative choice; releasing it one day earlier leads to the same conclusions (Section~\ref{sec:reeval-robust}).

\textbf{The protocol is the most favorable correct setting for past labels.} It releases each label at the end of its recovered window, the earliest time at which the label could have been computed (on tgbn-genre, up to one day later, which does not change our conclusions; Section~\ref{sec:reeval-robust}), whereas deployed systems typically receive labels later. Since label-based scores fall as the delay grows (Section~\ref{sec:reeval-robust}), our leakage-free scores of label-based methods are optimistic, and with edges available as before, a later release would only widen every advantage of edges over past labels that we report.

\textbf{Method coverage.} Among learned methods, we evaluate only NAVIS, in both of its modes. 
The leakage does not depend on the model, but how much a method gains from it does: on tgbn-trade, edge aggregation reaches near-perfect scores, whereas TGN gains only about 0.006~\cite{tgmparity}. Our conclusions about the two inputs rest on the blends and NAVIS, each compared with the model held fixed; other architectures may draw on the two inputs differently.

\textbf{Training and selection of NAVIS.} We use the released hyperparameters of NAVIS, chosen under the original protocol, without retuning them; retuning could raise its leakage-free scores somewhat but cannot restore the withheld information, whereas the blends are refit under each protocol, an asymmetry that may favor them. 

\section{Related Work}
\label{sec:related}

\textbf{Evaluation of temporal graph learning.} Critiques of temporal graph evaluation have so far focused mainly on link prediction. Poursafaei et al.\ showed that EdgeBank, a pure memorization baseline, performs strongly because the negative edges used in evaluation are easy, and proposed harder negative sampling~\cite{poursafaei2022towards}. TGB-Seq~\cite{yi2025tgbseq} curates datasets with few repeated edges to test whether models capture sequential dynamics, DyGLib~\cite{yu2023towards} and Yang et al.~\cite{yang2025towards} compare TGNNs under unified pipelines, and Lampert et al.\ analyze how batch-based evaluation distorts link prediction~\cite{lampert2026link}. For node affinity prediction, TGB and NAVIS report that heuristics built on past labels outperform TGNNs~\cite{huang2023temporal, feldman2026revisiting}, but, to our knowledge, when these labels become available has not been examined; the known issues of the TGB loop concern its edges (Table~\ref{tab:audit}).

\textbf{Leakage in machine learning evaluation.} Kaufman et al.\ formalize leakage in data mining~\cite{kaufman2011leakage}, Kapoor and Narayanan catalogue it across scientific fields~\cite{kapoor2023leakage}, and Yang et al.\ detect common forms of it in data science code by static analysis~\cite{yang2022data}. In recommender systems, data splits that ignore the global timeline let models learn from interactions that are not yet available at the time of a test instance, so that the relative order of models depends on how much future data leaks~\cite{ji2023critical}. These works concern which data a model may use; in streaming evaluation, leakage can also arise from when test labels are fed back as inputs, which Definition~\ref{def:pit} captures through the availability time (Sections~\ref{sec:pit} and~\ref{sec:lessons-dm}).

\textbf{Benchmark audits.} Flaws in a benchmark, rather than in the methods, can drive reported progress. Wu and Keogh find that many exemplars of popular time series anomaly detection benchmarks are trivial to solve and argue that much of the reported progress may be illusory~\cite{wu2023current}. In data management, Jin et al.\ show that annotation errors in text-to-SQL benchmarks distort scores and leaderboard rankings~\cite{jin2026annotations}, Tolovski et al.\ analyze the benchmarking debt of TPCx-AI~\cite{tolovski2026tpcxai}, re-evaluations of entity alignment and time series forecasting find under unified settings that traditional methods remain competitive with neural ones~\cite{leone2022critical, qiu2024tfb}, and Wang et al.\ find learned cardinality estimators more accurate than traditional ones but not yet ready for deployment~\cite{wang2021ready}. Our study belongs to this line of work: as for text-to-SQL, correcting the flaw changes both scores and rankings (Section~\ref{sec:reeval}). Unlike these flaws, ours are temporal: they concern when information reaches a model relative to its queries, which neither the labels nor the metric reveal on their own.

\section{Conclusion}
\label{sec:conclusion}

None of the four TGB node affinity datasets is evaluated in a point-in-time correct way: on the daily datasets, each label summarizes a single day about five days after its timestamp yet is fed back immediately, and on tgbn-trade, the edges from which each label is computed are processed before it is predicted. Without this leakage, label-based methods lose 3\% to 17\% on the daily datasets and edge aggregation about 14\% on tgbn-trade, the advantage of past labels over edges nearly vanishes or reverses, the best configuration of the state of the art and the rankings change, and a blend of training-free long-memory statistics matches or exceeds the state of the art under both protocols. Benchmarks should therefore store the availability time of each label and use a reference loop that respects it; our tools carry over to any benchmark that pairs an event stream with window-aggregated labels. Re-evaluating TGNNs and designing targets that reward fine-grained temporal modeling remain future work.


\section*{Code Availability}
The audit code is available at \url{https://github.com/quzah521/tgbn-leakage-audit-arxiv}.

\bibliographystyle{ACM-Reference-Format}
\bibliography{references}

\end{document}